\documentclass[conference]{IEEEtran}
\IEEEoverridecommandlockouts
\usepackage{cite}
\usepackage{amsmath,amssymb,amsfonts}
\usepackage{algorithmic}
\usepackage{graphicx}
\usepackage{textcomp}
\usepackage{xcolor}
\def\BibTeX{{\rm B\kern-.05em{\sc i\kern-.025em b}\kern-.08em
    T\kern-.1667em\lower.7ex\hbox{E}\kern-.125emX}}

\usepackage{algorithm2e}
\usepackage{array}
\usepackage{booktabs}
\usepackage{siunitx}
\usepackage{amsfonts}
\usepackage{hyperref}
\newcommand{\method}{DFBF}
\newcommand\mapvoc{mAP$^{0.5}$}
\newcommand\mapcoco{mAP$^{0.5:0.95}$}

\begin{document}
\bstctlcite{IEEEexample:BSTcontrol}

\title{DATA-FREE BACKBONE FINE-TUNING FOR PRUNED NEURAL NETWORKS
\thanks{Part of the work was supported by INTUITIVER (7547.223-3/4/), funded by the State Ministry of Baden-Württemberg for Sciences, Research and Arts and the State Ministry of Transport Baden-Württemberg.}
}

\author{\IEEEauthorblockN{Adrian Holzbock\IEEEauthorrefmark{1}, Achyut Hegde\IEEEauthorrefmark{2}, Klaus Dietmayer\IEEEauthorrefmark{1} and Vasileios Belagiannis\IEEEauthorrefmark{3}}
\IEEEauthorblockA{\IEEEauthorrefmark{1}\textit{Institute for Measurement-, Control-, and Microtechnology} \\
\textit{Ulm University}, Germany\\
adrian.holzbock@uni-ulm.de, klaus.dietmayer@uni-ulm.de}
\IEEEauthorblockA{\IEEEauthorrefmark{2}\textit{Institute of Information Security and Dependability} \\
\centerline{\textit{Karlsruhe Institute of Technology}, Germany}\\
achyut.hegde@kit.edu}
\IEEEauthorblockA{\IEEEauthorrefmark{3}\textit{Chair of Multimedia Communications and Signal Processing} \\
\centerline{\textit{Friedrich-Alexander-Universität Erlangen-Nürnberg}, Germany}\\
vasileios.belagiannis@fau.de}
}

\newcolumntype{C}[1]{>{\centering\arraybackslash}m{#1}}
\newcolumntype{L}[1]{>{\raggedright\arraybackslash}p{#1}}

\maketitle
\thispagestyle{empty}
\pagestyle{empty}

\begin{abstract}
Model compression techniques reduce the computational load and memory consumption of deep neural networks. After the compression operation, e.g. parameter pruning, the model is normally fine-tuned on the original training dataset to recover from the performance drop caused by compression. However, the training data is not always available due to privacy issues or other factors. In this work, we present a data-free fine-tuning approach for pruning the backbone of deep neural networks. In particular, the pruned network backbone is trained with synthetically generated images, and our proposed intermediate supervision to mimic the unpruned backbone's output feature map. Afterwards, the pruned backbone can be combined with the original network head to make predictions. We generate synthetic images by back-propagating gradients to noise images while relying on L1-pruning for the backbone pruning. In our experiments, we show that our approach is task-independent due to pruning only the backbone. By evaluating our approach on 2D human pose estimation, object detection, and image classification, we demonstrate promising performance compared to the unpruned model. Our code is available at \url{https://github.com/holzbock/dfbf}.
\end{abstract}

\begin{IEEEkeywords}
Data-Free Neural Network Pruning, Data-Free Object Detection Pruning, Data-Free Pose Estimation Pruning
\end{IEEEkeywords}

\section{Introduction}
\label{sec:intro}
Computer vision tasks like object detection~\cite{ren2015faster,9747512} or human pose estimation~\cite{kreiss2021openpifpaf,belagiannis2014holistic} have long been studied in the literature. Their leading performance though comes at the cost of high computational and memory requirements, making them difficult to deploy on resource-constrained devices. For deep neural networks, the compute and memory demands are often reduced with model compression, e.g.~parameter pruning~\cite{blalock2020state} or knowledge distillation~\cite{gou2021knowledge}. However, most neural network compression techniques normally require the availability of the training set. Due to privacy issues or large dataset size, access to the training dataset cannot always be guaranteed.

\begin{figure}[]
  \centering
  \includegraphics[width=\linewidth]{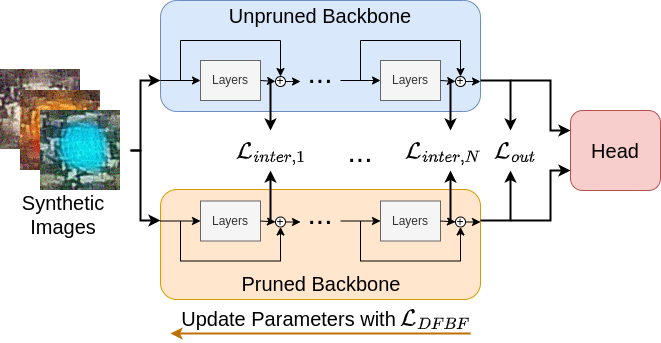}
  \caption{Training of the pruned backbone with label-free synthetic images. Besides the backbone's output feature map, some intermediate feature maps are used to recover the backbone's knowledge.}
  \label{img:loss_overview}
\end{figure}

As an alternative to the data-driven model compression~\cite{blalock2020state,gou2021knowledge,letaifa2022transformer}, approaches using only a few samples of the training dataset~\cite{9746024} and data-free methods have been developed to overcome the limited access to the original training dataset. The data-free approaches generate synthetic images to regain the knowledge after pruning~\cite{tang2021data} or transfer the knowledge with knowledge distillation~\cite{choi2020data} to a smaller model. The synthetic images can be generated with an additional generator network trained with the original model's knowledge~\cite{chen2019data,tang2021data}. Another method is to optimize noisy images by back-propagating directly onto the pixels~\cite{li2021mixmix}. In this work, we focus on neural network pruning similar to~\cite{tang2021data}. However, we rely on the back-propagation of gradients for image generation instead of employing a generator network. In contrast to existing methods, our approach is not limited to a specific task, but is applicable to different computer vision tasks. We achieve task-independence by only pruning the backbone of the model, keeping the task-dependent head unchanged.

We present a \textbf{D}ata-\textbf{F}ree \textbf{B}ackbone \textbf{F}ine-tuning approach (\method) for pruning the backbone of deep neural networks. In particular, we train the pruned backbone with synthetic images to mimic the output of the unpruned backbone. Later, we can use the pruned backbone with the original network head to make predictions. The synthetic images are generated by back-propagating gradients directly on noise images, while the backbone is pruned by $\ell_1$-pruning, leaving the parameters of the network head unmodified. Only backbone pruning keeps the pruning approach task-independent, as no training of the task-specific head is required. For the training of the backbone, we introduce an intermediate loss function that adjusts the pruned backbone's intermediate feature maps and output to match it with the original backbone. An overview of the proposed method is given in Fig.~\ref{img:loss_overview}. After the training, the original network's head can be attached to the pruned backbone for the task-specific output. 

We examine our approach on three vision tasks, namely object detection, human pose estimation, and image classification. First, we use the Faster R-CNN object detector~\cite{ren2015faster} for object detection and present the results for different pruning ratios on the COCO detection dataset~\cite{lin2014microsoft} and Pascal VOC dataset~\cite{Everingham10}. We evaluate pose estimation with the OpenPifPaf pose estimator~\cite{kreiss2021openpifpaf} trained on the COCO keypoint dataset~\cite{lin2014microsoft}. Additionally, we compare the performance on CIFAR10~\cite{krizhevsky2009learning} for image classification. As backbone, we use ResNet~\cite{he2016deep} and VGG~\cite{Simonyan15} models in our experiments.
To the best of our knowledge, we are the first to introduce a data-free fine-tuning method for pruning that can be applied to complex computer vision tasks like object detection and human pose estimation. The proposed intermediate loss function that applies the prediction of the unpruned model as ground truth enables the task independence of \method.

\section{Method}
\label{sec:method}
In the following, we present our data-free fine-tuning method for pruned neural networks, which process input images $\mathbf{x} \in \mathbb{R}^{3 \times W\times H}$, where $W$ and $H$ are the width and height, respectively. We assume that the neural network $\mathbf{y} = f(\mathbf{x}; \theta)$ can be divided into a backbone $\mathbf{z} = b(\mathbf{x}; \theta_b)$ and a head $\mathbf{y} = h(\mathbf{z}; \theta_h)$, where the head $h$ is a task-specific output network. $\mathbf{y}$ is the output prediction, $\mathbf{z}$ is the output feature map from the backbone forwarded to the task-specific head, while $\theta$, $\theta_b$, and $\theta_h$ are the model parameters of the whole model, the backbone, and the head, respectively. Since we only prune the model backbone, our approach is independent of the shape of the model prediction $\mathbf{y}$ and thus can be applied to different computer vision tasks.

\subsection{Preliminaries}
\label{subsec:prelim}

\paragraph*{Pruning}
Utilizing pruning algorithms reduces neural networks' resource demand by removing unnecessary or redundant parameters. Pruning convolutional neural networks can be divided into different groups. The parameter pruning only removes single weights~\cite{10.5555/2969239.2969366}, leading to sparse kernels and no satisfying reduction in computation. In contrast, filter pruning reduces the computation effort by removing entire filters~\cite{zhang2021akecp}. To decide which filters are pruned, the methods rely on the knowledge of the model, like in $\ell_1$-pruning~\cite{li2016pruning} and Batch Normalization pruning~\cite{liu2017learning}, or even use additional neural network layers~\cite{9746346}. Because of its simplicity and the fact that no extra neural network layers are needed, we rely on the $\ell_1$-pruning \cite{li2016pruning} to get the pruned backbone $b_p(\cdot)$ with its reduced parameters $\theta_{bp}$. However, every other filter pruning approach can be used to prune the model's backbone for \method. The $\ell_1$-pruning calculates for each filter $\mathcal{F}_i$ of a neural network layer the sum of the filter weights $w_i$ by $s_i = \sum|w_i|$ and removes filters with the smallest sum $s_i$ according to the required sparsity. The reduction of the current layer's filters makes it necessary to remove the corresponding input channels of the following layer. Furthermore, the pruning sparsity can be adapted to the number of filters in the layer, i.e. in layers with many filters, percentual more filters are pruned than in layers with fewer filters.

\paragraph*{Image Synthesis}
Since the original training set is unavailable, we synthesize training images by transferring knowledge from the model to a noisy image, similar to DeepInversion~\cite{yin2020dreaming}, but without being limited to the classification task. At the beginning of the image generation, a noise image $\mathbf{\hat{x}} \in \mathcal{R}^{3 \times w \times h}$ with width $w$ and height $h$ is fed into the backbone, while the synthetic image size can differ from the original $(W \neq w; H \neq h)$. Then, the loss $\mathcal{L}_{image}$ is propagated back onto the noise image $\mathbf{\hat{x}}$ to adapt the pixels directly without changing the model parameters. The loss $\mathcal{L}_{image}$ for optimizing the noise image independent of the underlying task is the weighted sum of the following three parts. The Batch Normalization loss is calculated between the batch statistics of the Batch Normalization layers and the statistics of the actual noise images. The total variance loss is defined by the $\ell_2$-norm of the differences between horizontally and vertically adjacent image pixels and is calculated on the noise image directly. Additionally, the loss is regularized by penalizing the $\ell_2$-norm of the entire image. The task independence of $\mathcal{L}_{image}$ is reached by not using the task-specific outputs of the head, whereby images of different computer vision tasks can be synthesized. To train the pruned model, we generate a synthetic dataset $\mathcal{D}$ containing $M$ synthetic images $\mathbf{\hat{x}}$.

\subsection{Overall Pruning Procedure}
\label{subsec:overall_procedure}
Our method is designed for pruning a neural network of any image-dependent task in a data-free manner. Our main contribution is the fine-tuning step after the pruning without relying on training data. Therefore, we focus on reducing the size of the backbone $b(\cdot)$ and keeping the head $h(\cdot)$ as is. We perform the task-independent data-free pruning in the following three steps: 1) image generation with an adapted loss function, 2) network pruning with the $\ell_1$-pruning, and 3) fine-tuning only the backbone with the proposed intermediate loss function. A systematic overview of the data-free fine-tuning of the backbone is given in Fig.\ref{img:loss_overview}. 

\subsection{Data-Free Training}
\label{subsec:dff}
During the pruning process, the model loses some of its knowledge which can be recovered in the following training procedure. In data-driven pruning methods, fine-tuning is performed with the original training dataset. In contrast, we use the synthetic dataset $\mathcal{D}$, which only contains pseudo images $\mathbf{\hat{x}}$, but no labels. However, the loss calculation in the standard training with the original data needs labels for the parameter update. To overcome the issue of having no ground truth labels for the pseudo images, we introduce a method that generates pseudo labels with the help of the unpruned model and applies them to improve the performance of the pruned model.

The final output of a neural network depends on the defined model task. Therefore, we propose not handling the output of the unpruned network's head as pseudo ground truth but instead using the output feature map $\mathbf{z}$ of the unpruned model's backbone. This design choice makes our approach independent of the task-specific model heads. More precisely, we define the loss as the $\ell_1$-loss between the output feature map of the unpruned backbone $\mathbf{z}$ and the pruned backbone $\mathbf{\hat{z}}$. Our output feature map loss $\mathcal{L}_{out}$ can be formulated as
\begin{equation}
    \mathcal{L}_{out} = \frac{1}{w_o * h_o} \sum_{i=0}^{w_o} \sum_{j=0}^{h_o} | \mathbf{\hat{z}}_{i,j} - \mathbf{z}_{i,j}|,
\end{equation}

where $w_o$ and $h_o$ are the width and height of the output feature map, respectively.

Additionally, we define an intermediate feature map loss $\mathcal{L}_{inter}$. For $\mathcal{L}_{inter}$, we calculate the $\ell_1$-loss between intermediate feature maps of the unpruned model $\mathbf{a}_n, n \in [ 1\dots N]$ and the corresponding feature maps of the pruned model $\mathbf{\hat{a}}_n, n \in [ 1\dots N]$, where $N$ defines the number of intermediate feature maps. Empirically, we found that the intermediate feature maps after the Batch Normalization layers~\cite{ioffe2015batch} lead to the best results. Since the feature maps behind the pruned layers differ in their dimensions from the unpruned model ones, it is impossible to calculate the loss between them. Therefore, we propose to skip single layers in the pruning and use them later for the loss calculation. Using $N$ intermediate feature maps for the loss calculation, $\mathcal{L}_{inter}$ can be calculated as 
\begin{equation}
    \mathcal{L}_{inter} = \sum_{n=1}^{N} \bigg( \frac{\mu_n}{w_n * h_n} \sum_{i=0}^{w_n} \sum_{j=0}^{h_n} | \mathbf{\hat{a}}_{n,i,j} - \mathbf{a}_{n,i,j}| \bigg).
\end{equation}
The influence of the different feature maps on the loss $\mathcal{L}_{inter}$ is defined by the parameter $\mu_n$, which we define as $\frac{n}{N + 1} * \gamma + 1$, where the parameter $\gamma$ is a scaling factor. $w_n$ and $h_n$ are the width and height of the intermediate feature map $n$. 

The overall loss $\mathcal{L}_{\method}$ in our data-free training combines $\mathcal{L}_{out}$ and $\mathcal{L}_{inter}$:
\begin{equation}
\label{eq: L_tadf}
    \mathcal{L}_{\method} = \mu_{out} \mathcal{L}_{out} + \mathcal{L}_{inter},
\end{equation}
where $\mu_{out}$ is the weighting factor of the output loss that we define as $\gamma + 1$. Importantly, $\mathcal{L}_{\method}$ is applied to optimize the parameters $\theta_{bp}$ of the pruned backbone $b_p(\cdot)$ and not to update the parameters $\theta_{h}$ of the head $h(\cdot)$.

During inference, we combine the optimized pruned backbone $\mathbf{\hat{z}} = b_p(\mathbf{x}, \theta_{bp})$ with the pre-trained head $\mathbf{\hat{y}} = h(\mathbf{\hat{z}}, \theta_h)$ to obtain the pruned model as:
\begin{equation}
\label{eq: predict}
\mathbf{\hat{y}} = h(b_p(\mathbf{x}, \theta_{bp}), \theta_h).
\end{equation}
For the prediction, we feed original images $\mathbf{x}$ to the model and get a slightly different model output $\mathbf{\hat{y}}$ because of the modified backbone weights $\theta_{p}$. Due to the backbone training with the synthetic images, the output of the original model $\mathbf{y}$ and the pruned model $\mathbf{\hat{y}}$ are nearly identical.

\begin{figure}[]
  \centering
  \includegraphics[width=0.9\linewidth]{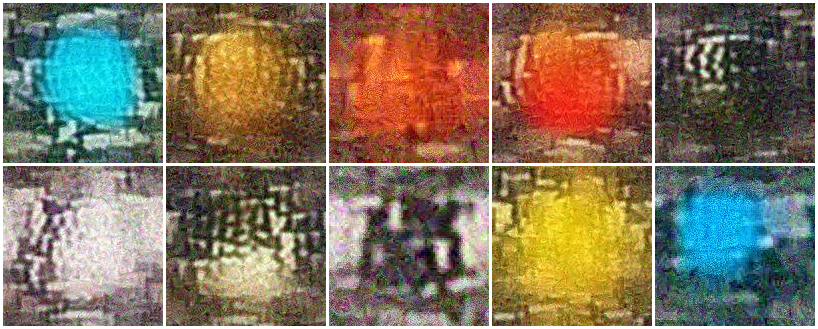}
  \caption{Synthetic images generated with Faster R-CNN object detector~\cite{ren2015faster} with a ResNet50~\cite{he2016deep} backbone trained on the COCO detection dataset~\cite{lin2014microsoft}.}
  \label{img:synthetic_images_detection}
\end{figure}

\begin{figure}[]
  \centering
  \includegraphics[width=0.9\linewidth]{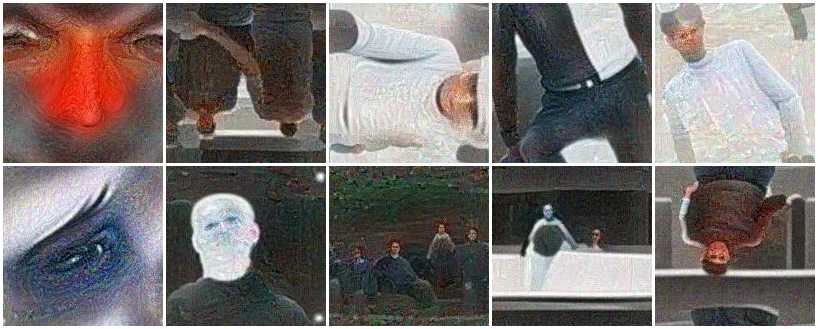}
  \caption{Synthetic images generated with an OpenPifPaf pose estimator~\cite{kreiss2021openpifpaf} with a ResNet50~\cite{he2016deep} backbone trained on the COCO keypoint dataset~\cite{lin2014microsoft}.}
  \label{img:synthetic_images_pose}
\end{figure}

\section{Experiments}
\label{sec:experiments}
We show the effectiveness of our proposed method in three challenging computer vision tasks: object detection, human pose estimation, and image classification. In object detection, we use Faster R-CNN~\cite{ren2015faster} trained on the COCO detection~\cite{lin2014microsoft} as well as on the Pascal VOC dataset~\cite{Everingham10}. The base for the human pose estimation is an Open PifPaf pose estimator~\cite{kreiss2021openpifpaf} trained on the COCO keypoint dataset~\cite{lin2014microsoft}. Both tasks utilize a ResNet50~\cite{he2016deep} backbone, and VGG16~\cite{Simonyan15} is used as an additional backbone in object detection. We prune between 10\% and 40\% of the backbone filters with $\ell_1$-pruning during the evaluation. In image classification, we compare with Tang et al.~\cite{tang2021data} and follow their evaluation protocol using Batch Normalization pruning~\cite{liu2017learning} for the VGG models~\cite{Simonyan15} and $\ell_1$ pruning~\cite{li2016pruning} for the ResNet models~\cite{he2016deep}. The pruned backbone is trained in each task with 1600 synthetic images using the SGD optimizer with a learning rate of 0.01, a momentum of 0.9, and a weight decay of \num{5e-4}.

\begin{table*}[]
  \caption{Results of \method\ for object detection when pruning Faster R-CNN~\cite{ren2015faster} with ResNet50~\cite{he2016deep} and VGG16~\cite{Simonyan15} backbone trained on COCO detection~\cite{lin2014microsoft} and Pascal VOC~\cite{Everingham10}. OpenPifPaf~\cite{kreiss2021openpifpaf} with a ResNet50~\cite{he2016deep} backbone trained on COCO keypoints~\cite{lin2014microsoft} for pose estimation. \textit{w/o fine-tuning}: without fine-tuning after pruning; \textit{orig. img}: original images and $\mathcal{L}_{\method}$ loss.}
  \label{tab:results_pruning}
  \begin{center}
  {\small{
  \begin{tabular}{lC{1.8cm}C{1.6cm}C{2.3cm}C{2.3cm}C{2.2cm}C{2.2cm}C{2.2cm}}
    \toprule
    Method & Removed Params in \% & Removed Filters in \% & COCO detection \mapcoco\ in \% & COCO detection \mapcoco\ in \% & VOC \newline \mapvoc\ in \% & COCO keypoints AP in \% \\
    \midrule
    Backbone & ResNet50/ VGG16 & ResNet50/ VGG16 & ResNet50 & VGG16 & ResNet50 & ResNet50 \\
    \midrule
    Baseline & 0/0 & 0 & 37.4 & 23.0 & 78.1 & 68.1 \\ 
    \midrule
    w/o fine-tuning & 14/16 & 10 & 34.0 & 19.4 & 74.7 & 57.9 \\
    orig. img & 14/16 & 10 & 35.8 & 21.3 & 77.7 & 63.3 \\ 
    \method & 14/16 & 10 & 36.2 & 20.9 & 77.6 & 65.2 \\ 
    \midrule
    w/o fine-tuning & 28/32 & 20 & 24.6 & 11.6 & 60.3 & 36.1 \\ 
    orig. img & 28/32 & 20 & 36.3 & 20.0 & 77.4 & 60.2 \\
    \method & 28/32 & 20 & 33.4 & 18.8 & 75.1 & 62.2 \\ 
    \midrule
    w/o fine-tuning & 40/46 & 30 & 13.8 & 5.3 & 36.9 & 20.4 \\ 
    orig. img & 40/46 & 30 & 35.3 & 18.1 & 76.0 & 55.5 \\ 
    \method & 40/46 & 30 & 28.0 & 15.2 & 71.5 & 55.5 \\ 
    \midrule
    w/o fine-tuning & 52/59 & 40 & 4.6 & 2.2 & 20.7 & 4.6 \\ 
    orig. img & 52/59 & 40 & 33.1 & 14.1 & 74.0 & 48.8 \\ 
    \method & 52/59 & 40 & 19.1 & 10.7 & 63.2 & 49.1 \\ 
  \bottomrule
\end{tabular}
}}
\end{center}
\end{table*}

To the best of our knowledge, we are the first to prune an object detection or a human pose estimation model in a data-free manner. Therefore, we cannot make a direct comparison with other methods. We compare \method\ with the unpruned baseline referred to as \textit{Baseline}, the pruned but not fine-tuned model referred to as \textit{w/o fine-tuning}, and the pruned model trained with 1600 random sampled original training images and our intermediate loss $\mathcal{L}_{\method}$ referred to as \textit{orig. img}. Additionally, we report the number of removed filters and parameters, while the parameter count differs from the filter count because of different pruning sparsities in separate layers.

\subsection{Object Detection}
\label{subsec:eval_od}
We set the synthetic image resolution for the object detection task to $250 \times 250$ pixels and $\gamma$ to 1. The results for pruning a Faster R-CNN object detection model~\cite{ren2015faster} in a data-free manner are shown in Tab.~\ref{tab:results_pruning}. For both datasets and backbones, the performance evolves similarly during pruning and training. The performance of the object detector after pruning depends on the pruning sparsity and decreases with higher rates. Also, we can see that with \method, the performance of the pruned model can recover during training with the synthetic images near the initial results. The model's performance trained with synthetic images is behind the model trained with the original images. Furthermore, the influence of synthetic images at higher pruning rates (30\% and 40\%) can be seen. Here, the difference between the model trained with the synthetic and the original images increases compared to lower pruning rates (10\% and 20\%). We show some synthetic images generated from an object detector trained on the COCO detection dataset~\cite{lin2014microsoft} in Fig.~\ref{img:synthetic_images_detection}. Compared to the original images, no objects can be recognized in the synthetic images. The abstract look of the synthetic images can cause the performance gap between the synthetic and original training image’s performance using high pruning sparsities.

\subsection{Human Pose Estimation}
\label{subsec:eval_pe}
Besides object detection, we evaluate \method\ for human pose estimation which is often the base for gesture recognition~\cite{holzbock2022spatio} or human motion prediction~\cite{ijcai2022p111}. We use synthetic images with a resolution of $160 \times 160$ and 6 as value for $\gamma$. The results for the human pose estimation are shown in Tab.~\ref{tab:results_pruning}. As for object detection, the performance decreases with a higher pruning rate in human pose estimation. For all pruning rates, \method\ can recover the performance after pruning. In contrast to object detection, we perform better on human pose estimation with the synthetic images than with the original training images. This could be due to the difference between the synthetic images obtained from both tasks. Synthetic images generated with OpenPifPaf trained on the COCO keypoint dataset can be seen in Fig.~\ref{img:synthetic_images_pose}. Compared to the images generated in the object detection task in Fig.~\ref{img:synthetic_images_detection}, the pose images show pose information and pose-related details such as faces. Therefore, the generated images' variation can be broader compared to the original images. Moreover, the synthetic images show only patterns important for human pose estimation, and the pruned model can concentrate on learning the essential shapes.

\subsection{Image Classification}
Unlike the other tasks, we compare image classification with the state-of-the-art approach from Tang et al.~\cite{tang2021data}. During pruning the image classification model, the resolution of the synthetic training images is $32 \times 32$, and we set the factor $\gamma$ to 0. Tab.~\ref{tab:results_img_class} presents the CIFAR10~\cite{krizhevsky2009learning} results for different ResNet and VGG models fine-tuned with \method\ and the approach of Tang et al.~\cite{tang2021data}. In the pruning, we remove 30\% and 50\% of the filters from the baseline models. The results show that our approach is on par with Tang et al.~\cite{tang2021data}, while we are not limited to the classification task due to the proposed task-agnostic loss function. 

\begin{table}[]
  \caption{Comparison of our method with Tang et al.~\cite{tang2021data} on the CIFAR10~\cite{krizhevsky2009learning} dataset with VGG16/19~\cite{Simonyan15} and ResNet18/34~\cite{he2016deep}. Model accuracy and removed (RM) filters are given in \%.}
  \label{tab:results_img_class}
  \begin{center}
  {\small{
  \begin{tabular}{lC{0.9cm}C{0.9cm}C{0.9cm}C{0.9cm}C{0.9cm}}
    \toprule
    Method & RM Filters & VGG 16 & VGG 19 & ResNet 18 & ResNet 34 \\
    \midrule
    Baseline & 0 & 94.00 & 93.95 & 93.07 & 93.33 \\
    \midrule
    \method & 30 & 92.85 & 92.85 & 92.66 & 92.92 \\ 
    Tang et al. & 30 & 92.78 & 92.82 & - & - \\
    \midrule
    \method & 50 & 92.07 & 92.14 & 92.48 & 92.60 \\ 
    Tang et al. & 50 & 92.43 & 92.78 & 92.68 & 93.25 \\
    \bottomrule
\end{tabular}
}}
\end{center}
\end{table}

\section{Ablation Studies}
\label{sec:ablations}
An essential part of \method\ is the proposed intermediate loss function $\mathcal{L}_{\method}$ for which we do further investigations. We use the OpenPifPaf settings from Sec.~\ref{sec:experiments} as standard settings.

In $\mathcal{L}_{\method}$, we take the feature map after each residual connection and the output feature map to calculate the loss and get an overall performance of 62.2\%. The performance drops by 1.0\% AP to 61.2\% AP when skipping every second intermediate feature map in the loss calculation. Neglecting all intermediate feature maps and using only the output feature map in the loss calculation decreases the performance to 44.7\% AP. These experiments demonstrate the importance of the intermediate feature maps on $\mathcal{L}_{\method}$. Furthermore, we present the effect of the scaling factor $\gamma$ in Tab.~\ref{tab:results_weighting_ablation}. We vary the impact of the different feature maps on $\mathcal{L}_{\method}$ by setting $\gamma \in [2, \dots, 8]$. With increasing $\gamma$, the overall performance improves.

\begin{table}[]
\caption{Results of different scaling factors $\gamma$ in $\mathcal{L}_{\method}$.}
\label{tab:results_weighting_ablation}
\begin{center}
{\small{  
\begin{tabular}{lccccccccc}
    \hline
    $\gamma$ & 
    2 & 3 & 4 & 5 & 6 & 7 & 8 \\
    \hline
    AP & 60.0 & 60.8 & 61.4 & 61.8 & 62.2 & 62.4 & 62.6 \\
  \hline
\end{tabular}
}}
\end{center}
\end{table}

\section{Conclusion}
\label{sec:conclusion}
We presented a data-free backbone fine-tuning approach for pruning the backbone of deep neural networks. Our approach relies on synthetically generated images to fine-tune the backbone of the pruned neural network, where pruning is based on the $\ell_1$ norm. Notably, we proposed an intermediate loss function to match the pruned backbone's output feature map such that the pruned backbone can later be combined with the original network head to perform predictions. Our evaluations showed that our approach is task-independent by evaluating the tasks of object detection, human pose estimation, and image classification.

\bibliographystyle{IEEEtran}
\bibliography{EUSIPCO_ref} 

\end{document}